\pdfoutput=1
\documentclass[]{interact}

\usepackage{xcolor}

\usepackage{epstopdf}
\usepackage{subfigure}

\usepackage{natbib}
\bibpunct[, ]{(}{)}{;}{a}{}{,}
\renewcommand\bibfont{\fontsize{10}{12}\selectfont}

\usepackage{graphicx}
\usepackage{amssymb}        
\usepackage{todonotes}
\usepackage{adjustbox}
\usepackage{float}
\usepackage{color,soul}
\usepackage{todonotes}
\usepackage{listings}
\usepackage{hyperref}
\usepackage{array}
\usepackage{makecell}
\usepackage{tabularx}

\usepackage[noabbrev, capitalise]{cleveref}

\restylefloat{table}

\definecolor{codebkg}{HTML}{EBEBEB}

\definecolor{cjmrozanec}{RGB}{0,0,139}
\definecolor{cinovalija}{RGB}{0,0,216}
\definecolor{cralonso}{RGB}{78,193,255}
\definecolor{cdreforgiato}{RGB}{1,166,255}
\definecolor{cncauli}{RGB}{0,117,180}
\definecolor{cisoldatos}{RGB}{223,207,50}
\definecolor{cibabis}{RGB}{223,178,50}
\definecolor{cdkyriazis}{RGB}{255,163,78}
\definecolor{cspyros}{RGB}{231,111,0}
\definecolor{cgsofianidis}{RGB}{205,98,0}
\definecolor{chtavakoli}{RGB}{0,142,12}
\definecolor{cssuh}{RGB}{0,91,8}
\definecolor{cvgezer}{RGB}{0,65,6}
\definecolor{cdpapamartzivanos}{RGB}{181,210,183}
\definecolor{ctgiannetsos}{RGB}{133,181,137}
\definecolor{csamenesidou}{RGB}{101,162,106}
\definecolor{ceveliou}{RGB}{90,148,95}

\newcommand{\orcidauthorJMR}{0000-0002-3665-639X}
\newcommand{\orcidauthorIN}{0000-0003-2598-0116}
\newcommand{\orcidauthorPZ}{0000-0002-6630-3106}
\newcommand{\orcidauthorKK}{0000-0002-4918-0650}

\newcommand{\orcidauthorSSuh}{0000-0003-3723-1980}
\newcommand{\orcidauthorEV}{0000-0001-9730-1720}
\newcommand{\orcidauthorDP}{0000-0002-9471-5415}
\newcommand{\orcidauthorTG}{0000-0003-0663-2263}
\newcommand{\orcidauthorSAM}{0000-0003-2446-5470}

\newcommand{\orcidauthorNC}{0000-0002-9611-0655}
\newcommand{\orcidauthorAM}{0000-0001-6768-4599}
\newcommand{\orcidauthorDRR}{0000-0001-8646-6183}
\newcommand{\orcidauthorDK}{0000-0001-7019-7214}
\newcommand{\orcidauthorGS}{0000-0002-9640-6317}

\newcommand{\orcidauthorJS}{0000-0002-6668-3911}
\newcommand{\orcidauthorDM}{0000-0002-0360-6505}
\newcommand{\orcidauthorBF}{0000-0002-8585-9388}

\begin{document}



\title{Human-Centric Artificial Intelligence Architecture for Industry 5.0 Applications}

\author{
\name{Jo\v{z}e M. Ro\v{z}anec*\textsuperscript{a,b,c}\thanks{Corresponding author: Jo\v{z}e M. Ro\v{z}anec. Email: joze.rozanec@ijs.si}\orcidauthorJMR{}, Inna Novalija\textsuperscript{b}\orcidauthorIN{}, Patrik Zajec\textsuperscript{a,b}\orcidauthorPZ{}, Klemen Kenda\textsuperscript{a,b,c}\orcidauthorKK{}, Hooman Tavakoli\textsuperscript{k}, Sungho Suh\textsuperscript{k}\orcidauthorSSuh{}, Entso Veliou\textsuperscript{e}\orcidauthorEV{}, Dimitrios Papamartzivanos\textsuperscript{d}\orcidauthorDP{}, Thanassis Giannetsos\textsuperscript{d}\orcidauthorTG{}, Sofia Anna Menesidou\textsuperscript{d}\orcidauthorSAM{}, Ruben Alonso\textsuperscript{f}, Nino Cauli\textsuperscript{f}\orcidauthorNC{}, Antonello Meloni\textsuperscript{g}\orcidauthorAM{}, Diego Reforgiato Recupero\textsuperscript{f,g}\orcidauthorDRR{}, Dimosthenis Kyriazis\textsuperscript{h}\orcidauthorDK{}, Georgios Sofianidis\textsuperscript{h}\orcidauthorGS{}, Spyros Theodoropoulos\textsuperscript{h,i}, Bla\v{z} Fortuna\textsuperscript{b,c}\orcidauthorBF{}, Dunja Mladeni\'{c}\textsuperscript{b}\orcidauthorDM{}, John Soldatos\textsuperscript{j}\orcidauthorJS{}}
\affil{\textsuperscript{a}Jo\v{z}ef Stefan International Postgraduate School, Jamova 39, 1000 Ljubljana, Slovenia; \textsuperscript{b}Jo\v{z}ef Stefan Institute, Jamova 39, 1000 Ljubljana, Slovenia; \textsuperscript{c}Qlector d.o.o., Rov\v{s}nikova 7, 1000 Ljubljana, Slovenia; \textsuperscript{d}Ubitech Ltd, Digital Security \& Trusted Computing Group, Athens, Greece; \textsuperscript{e}Department of Informatics and Computer Engineering, University of West Attica, Agiou Spyridonos Street, 12243, Egaleo, Athens, Greece; \textsuperscript{f}R2M Solution Srl, Pavia, Italy; \textsuperscript{g}Department of Computer Science, University of Cagliari, Cagliari, Italy; \textsuperscript{h}Department of Digital Systems, University of Piraeus, Piraeus, Greece; \textsuperscript{i}Department of Electrical and Computer Engineering, National Technical University of Athens, Athens, Greece; \textsuperscript{j}INTRASOFT International, 19.5 KM Markopoulou Ave., GR 19002 Peania, Greece; \textsuperscript{k}German Research Center for Artificial Intelligence (DFKI), 67663 Kaiserslautern, Germany}}

%

\maketitle

\begin{abstract}
Human-centricity is the core value behind the evolution of manufacturing towards Industry 5.0. Nevertheless, there is a lack of architecture that considers safety, trustworthiness, and human-centricity at its core. Therefore, we propose an architecture that integrates Artificial Intelligence (Active Learning, Forecasting, Explainable Artificial Intelligence), simulated reality, decision-making, and users' feedback, focusing on synergies between humans and machines. Furthermore, we align the proposed architecture with the Big Data Value Association Reference Architecture Model. Finally, we validate it on three use cases from real-world case studies.
\end{abstract}


\section{Introduction}\label{S:INTRODUCTION}
The development of new technologies and their increasing democratization have enabled the digitalization of the manufacturing domain. Furthermore, the digitalization and application of cutting-edge technologies have been fostered by many governments through national programs (e.g., \textit{Industrie 4.0}, \textit{Advanced Manufacturing Partnership}, or \textit{Made in China 2025} and others) as means to revitalize the industry and face societal changes such as the increasingly aging population (\cite{kuo2019industrial}). The Industry 4.0 paradigm, first presented in the Hannover Industrial Fair in 2011, aims to leverage the latest technologies (e.g., Internet of Things, cloud computing, and artificial intelligence, among others), and paradigms (e.g., Cyber-Physical Systems (\cite{lee2006cyber,rajkumar2010cyber,xu2019big}), and Digital Twins (\cite{grieves2015,grieves2017digital,tao2017digital})), to enable innovative manufacturing functionalities. Research regarding Industry 4.0 aims to devise means to realize mass customization, predictive maintenance, zero-defect manufacturing, and smart product lifecycles management (\cite{soldatos2019digital,lim2020state})). Furthermore, there is an increasing awareness regarding the complementarity of skills between humans and machines and the opportunity to foster human-centric solutions, which is one of the core principles in the emerging Industry 5.0.

When speaking about artificial intelligence, we must distinguish between General Artificial Intelligence (which aims to create machines capable of thinking and reasoning like humans), and Narrow Artificial Intelligence (which aims to solve specific problems automating specific and repetitive tasks). Industrial Artificial Intelligence can be considered a specific case of Narrow Artificial Intelligence, applied to industry. Artificial intelligence is being increasingly adopted in manufacturing, leading to work design, responsibilities, and dynamics changes. Artificial intelligence techniques can provide insights and partially or even fully automate specific tasks, while human input or decision-making remains critical in some instances. When insights are required for decision-making, it is of utmost importance to understand the models' rationale and inner workings to ensure the models can be trusted and responsible decisions made based on their outcomes (\cite{ahmed2022artificial}). Among the human-machine collaboration approaches, we find mutual learning, which considers learning to be a bidirectional process and a reciprocal collaboration between humans and machines when performing shared tasks (\cite{ansari2018autodidact,ansari2018rethinking}). Another possible approach is active learning, which assumes the machine learning model can learn from carefully selected data and leverage the knowledge and expertise of a human expert in a human-in-the-loop system. Furthermore, interactions between humans and machines can be enhanced by developing proper interfaces. For example, spoken dialog systems and voice-user interfaces attempt to do so by mimicking human conversations (\cite{mctear2016conversational,jentzsch2019conversational}). To ensure artificial intelligence systems are human-centered, much research is being invested in ensuring such systems remain secure and comply with ethical principles (\cite{shneiderman2020bridging}). Security challenges involve multiple aspects, such as ensuring that the integration between business and industrial networks remains secure (\cite{ani2017review}), and data-related aspects critical to artificial intelligence, such as protecting the Confidentiality, Integrity, and Availability (CIA) of data (\cite{wu2018cybersecurity,mahesh2020survey}). Compliance with ethical principles can be realized through three building blocks: (i) provide a framework of ethical values to support, underwrite and motivate (SUM) responsible data design and the use of the ecosystem, (ii) a set of actionable principles to ensure fairness, accountability, sustainability and transparency (FAST principles), and (iii) a process-based governance framework (PGB framework) to operationalize (i) and (ii) (\cite{leslie2019understanding}).

The richness of manufacturing use cases and the high level of shared challenges require standards to build a common ground to ensure the components' interoperability and that best practices are applied to develop and integrate them. Furthermore, there is a need to develop a unified architecture based on standards and reference architecture components to tackle the challenges described above. Among the reference architectures relevant to the field of manufacturing, we find the Reference Architecture Model for Industry 4.0 (RAMI 4.0 - presents the main building blocks of Industry 4.0 systems) (\cite{schweichhart2016reference}), the Industrial Internet Reference Architecture (IIRA - specifies a common architecture framework for interoperable IoT systems) (\cite{IIRA}), the Industrial Internet Security Framework (IISF) (\cite{IISF16}), and the Big Data Value Association (BDVA) Reference Architecture (\cite{renones2018european}). While the RAMI 4.0 and IIRA do not address the security and safety aspects, these are addressed by the ISSF and BDVA. Furthermore, the BDVA reference architecture provides structure and guidelines to structure big data software, foster data sharing, and enable the use of artificial intelligence in its components while ensuring compliance with standards. The requirements that emerge from the application of artificial intelligence have also crystallized in specific architectures. \cite{spelt1991hybrid} describes how machine learning models and expert systems can be combined, to leverage their complimentary strengths. \cite{xu2020edge} surveys multiple authors on how they realize edge intelligence, considering three key components: data, model, and computation. In the same line, \cite{yang2019federated} surveys authors on federated learning, considering horizontal and vertical federated learning, and federated transfer learning, while describing architectures that implement them. A slightly different approach is considered by \cite{wan2020artificial}, who describe an architecture combining cloud computing, edge computing, and local computing paradigms. The authors consider four major architecture components: smart devices, smart interaction, an artificial intelligence layer, and smart services.

In this work, we evolve and detail the architecture proposed in \cite{rovzanec2021stardom}, which addresses the void of an architecture specification that tackles the needs of trusted and secure artificial intelligence systems in manufacturing, seeking human-machine synergies by considering humans-in-the-loop. The human at the center of the manufacturing evolution represents the core of the evolution towards Industry 5.0 (\cite{nahavandi2019industry,EC2020}). Furthermore, we map the proposed architecture modules to the BDVA reference architecture and ISSF framework to ensure compatibility and show how their complimentary views can coincide in a single solution.

The rest of this paper is structured as follows: Section~\ref{S:RELATED-WORK} presents related work, and Section~\ref{S:REFERENCE-ARCHITECTURE} introduces three values-based principles and describes the proposed architecture. Next, section \ref{S:USE-CASES-OVERVIEW} describes the validating use cases, while Section \ref{S:EXPERIMENTS-AND-RESULTS} describes the experiments we conducted and the results we obtained. Finally, in Section~\ref{S:CONCLUSION}, we provide our conclusions and outline future work.

\section{Related Work}\label{S:RELATED-WORK}

\subsection{Industry 5.0}
Industry 4.0 was introduced at the Hannover Trade Fair in 2011, aiming to introduce new technologies into manufacturing with the purpose of achieving high levels of operational efficiency and productivity (\cite{sanchez2020industry}). While technology is emphasized as a means toward greater efficiency and productivity to enhance competitiveness in the global market, some emphasis have been placed on using such technology to reach certain level of human-centricity through the concept of Operator 4.0. Operators 4.0 are operators who will be assisted by systems providing relief from physical and mental stress, without compromising the production objectives (\cite{romero2016towards,romero2020operator,kaasinen2020empowering}). Industry 5.0 is envisioned as a co-existing industrial revolution (\cite{xu2021industry}), for which two visions have emerged: (i) one that refers to human-robot co-working, and (ii) a second one as a bioeconomy where renewable biological resources are used to transform existing industries (\cite{demir2019industry}). In this work, we focus on Industry 5.0 as a value-driven manufacturing paradigm and revolution that highlights the importance of research and innovation to support the industry while placing the well-being of the worker at the center of the production process (\cite{xu2021industry}). Such a revolution must attempt to satisfy the needs placed in the Industrial Human Needs Pyramid, which range from workplace safety to the development of a trustworthy relationship between humans and machines that enables the highest level of self-esteem and self-actualization, realizing and fulfilling their potential (\cite{lu2022outlook}). It aims to intertwine machines and humans in a synergistic collaboration to increase productivity in the manufacturing industry while retaining human workers. Furthermore, it seeks to develop means that enable humans to unleash their critical thinking, creativity, and domain knowledge. At the same time, the machines can be trusted to autonomously assist on repetitive tasks with high efficiency, anticipating the goals and expectations of the human operator, and leading to reduced waste and costs (\cite{nahavandi2019industry,demir2019industry,maddikunta2021industry}). Such communication and collaborative intelligence enable the development of trustworthy coevolutionary relationships between humans and machines. To foster the development of trustworthy coevolutionary relationships, interfaces must consider the employee's characteristics (e.g., age, gender, and level of education, among others) and the organizational goals. One example of collaboration between humans and machines is realized with cobots, where the cobots share the same physical space, sense and understand the human presence, and can perform tasks either independently, simultaneously, sequentially, or in a supportive way (\cite{el2019cobot}). 

In order to realize the Industry 5.0 vision, the focus must be shifted from individual technologies to a systematic approach rethinking how to (a) combine the strengths of humans and machines, (b) create digital twins of entire systems, and (c) widespread use artificial intelligence, with a particular emphasis generation of actionable items for humans. While research regarding Industry 5.0 is incipient, it has been formally encouraged by the European Commission through a formal document released back in 2021 (\cite{Industry50}).

\subsection{Considering standards and regulations}\label{STANDARDS OVERVIEW}
In order to realize the vision laid out for Industry 5.0, constraints and directions imposed by existing regulations must be considered. Furthermore, standards should be taken into account to ensure that the fundamental blocks can be universally understood and adopted to achieve compatibility and interoperability. 

Cybersecurity is considered a transversal concern in the architecture presented in this work. Among the standards and regulations that relate to it we must mention the ISO 27000 family of standards (\cite{ISO27000}), the USA Cybersecurity Information Sharing Act (CISA) (\cite{CISA2015}), the EU Cybersecurity Act (\cite{cybersecurityAct}), and the EU Network and Information Security Directive II (NIS II) (\cite{NISII}). The ISO 27000 standards defined a common vocabulary and provided an overview of information security management systems. The NIS II directive aimed to force certain entities and sectors of the European Union to take measures to increase the overall cybersecurity level in Europe. The European Union Cybersecurity Act provided complementary legislation by establishing a cybersecurity certification framework for products and services and granted a permanent mandate to the EU agency for cybersecurity (ENISA) to inform the public regarding certification schemas and issue the corresponding certificates. Finally, the CISA established a legal ground for information sharing between the USA government agencies and non-government entities for cyberattack investigations.

When considering data management, much emphasis is being put on privacy. The General Data Protection Regulation (GDPR) (\cite{GDPR}), ePrivacy directive (\cite{ePrivacy}), or Data Governance Act (\cite{dataGovernanceAct}), issued by the European Union, are relevant when managing and sharing data, especially personal or sensitive data. The GDPR establishes a legal framework setting guidelines to process and collect personal information of persons living in the European Union. The ePrivacy directive regulates data protection and privacy, emphasizing issues related to confidentiality of information, treatment of spam, cookies, and traffic data. Finally, the Data Governance Act promotes wider re-use of data, using secure processing environments, data anonymization techniques (e.g., differential privacy), and synthetic data creation; and establishes a licensing regime for data intermediaries between data holders and data users. While these regulations and directives must be considered, we provide no systemic solution from an architectural point of view.

Finally, given the increasing adoption of artificial intelligence, a legislative effort is being made to regulate its use. For example, the Artificial Intelligence Act (\cite{AIACT}), issued in the European Union, was the first law of this kind issued by a significant regulator worldwide. The law categorizes artificial intelligence applications into three risk categories: (a) unacceptable risk (e.g., social scoring systems), which are banned, (b) high-risk (e.g., resume scanning applications), which are subject to specific legal requirements, and (c) applications that do not fall into categories (a) and (b), which remain unregulated. Another example is a law issued by the Federative Republic of Brazil (\cite{AIACTBR}), which establishes the principles, obligations, rights, and governance instruments regarding the use of artificial intelligence.

While the abovementioned list is not exhaustive, it provides a high-level view of the main concerns and topics that must be considered.

\subsection{Enabling technologies}
In order to realize a human-centric artificial intelligence architecture for Industry 5.0 applications, a set of technologies that enable a human-centric approach we consider a set of technologies must be taken into account. We consider five of them related to the field of artificial intelligence: (i) active learning, (ii) explainable artificial intelligence, (iii) simulated reality, (iv) conversational interfaces, and (v) security. Below we introduce some related work regarding each of them, and in Section \ref{S:REFERENCE-ARCHITECTURE} describe the corresponding architecture building blocks.

\subsubsection{Active learning}
The adoption of artificial intelligence in manufacturing and the complementarity of the machine and human capabilities is reshaping jobs, and human-machine cooperation opportunities are emerging. One way to realize such human-machine cooperation is through the Active Learning paradigm, which considers an artificial intelligence model can be improved by carefully selecting a small number of data instances to satisfy a learning objective (\cite{settles2009active}). Active Learning is built upon three assumptions: (i) the learner (artificial intelligence model) can learn by asking questions (e.g., request a target variable's data), (ii) there is an abundance of questions that can be asked (e.g., data, either gathered or synthetically created, without a target value), and (iii) there is a constrained capacity to answer such questions (and therefore, the questions must be carefully selected) (\cite{elahi2016survey}). Therefore, applied research is focused on how to structure use case solutions so that through a human-in-the-loop, artificial intelligence models can benefit from human expertise to make decisions and provide valuable input, which is later used to enhance the models (\cite{kumar2020active,schroder2020survey,budd2021survey}).

We discriminate between data obtained from real sources and synthetic data (created through some procedure) regarding the source of the data. Synthetic data is frequently used to enlarge the existing data or to generate instances that satisfy specific requirements when similar data is expensive to obtain. While many techniques and heuristics have been applied in the past to generate synthetic data, the use of Generative Adversarial Networks (GANs) has shown promising results and been intensely researched \cite{zhu2017generative,mahapatra2018efficient,sinha2019variational,mayer2020adversarial}). Strategies related to data selection are conditioned by how data is generated and served. If the data is stored, data instances can be scanned and compared, and some latency can be tolerated to make a decision. On the other hand, decisions must be made at low latency in a streaming setting, and the knowledge is constrained to previously seen instances. Data selection approaches must consider informativeness (quantifying the uncertainty associated to a given instance, or the expected model change), representativeness (number of samples similar to the target sample), or diversity criteria (selected samples scatter across the whole input space) (\cite{wu2018pool}). Popular approaches for classification problems are the random sampling, query-by-committee (\cite{seung1992query}), minimization of the Fisher information ratio (\cite{padmanabhan2014active}), or hinted sampling with Support Vector Machines (\cite{li2015active}).

Active learning has been applied to several manufacturing use cases. Nevertheless, applied research in the manufacturing sector remains scarce (\cite{samsonov2019more,meng2020machine}), but its relevance increases along with the proliferation of digital data and democratization of artificial intelligence. In the scientific literature, authors report using Active Learning to tackle quality control, predictive modeling, and demand forecasting. For example, active learning for quality control was applied to predict the local displacement between two layers on a chip (\cite{dai2018towards}) or gather users' input in visual quality inspection of printed company logos on the manufactured products (\cite{trajkova2021active}). In predictive modeling, it was applied in the aerospace industry to assist a model in predicting the shape control of a composite fuselage (\cite{yue2020active}). Finally, in the demand forecasting use case, the authors explored using active learning to recommend media news and broaden the logisticians' understanding of the domain while informing relevant events that could affect the demand, to reach better decisions (\cite{zajec2021help}). Regardless of the successful application in several manufacturing use cases, active learning is not widely adopted in manufacturing and could be applied to enhance cybersecurity capabilities and fatigue monitoring systems (\cite{li2019fatigue}), among others.

\subsubsection{Explainable Artificial Intelligence}
While artificial intelligence was applied to manufacturing problems in the past (\cite{bullers1980artificial}), it has become increasingly common to rely on artificial intelligence models to automate certain tasks and provide data-based insights (\cite{chien2020artificial}). When human decision-making relies on artificial intelligence models' outcomes, enough information regarding the models' rationale for such a forecast must be provided (\cite{ribeiro2016should,lundberg2017unified}). Such information enables the user to assess the trustworthiness and soundness of the provided forecast and therefore ensure decisions are made responsibly (\cite{das2020opportunities}). Research on techniques and approaches that convey information regarding the rationale behind the artificial intelligence models, or an approximation to it, and how such information is best presented to the users, is done in a sub-field of artificial intelligence, known as eXplainable Artificial Intelligence (XAI) (\cite{henin2021multi}). Such approaches can be classified according to different taxonomies. Among them, there is consensus that artificial intelligence models can be considered either \textit{white-box models} (inherently interpretable models), or \textit{black-box models} (models that remain opaque to the users) (\cite{loyola2019black}). Regarding the characteristics of the explanation, \cite{angelov2021explainable} divide XAI methods into four groups, considering whether (i) the explanations are provided at a local (for a specific forecast) or global (for the whole model) level, (ii) the models are transparent or opaque to the users, (iii) the explainability techniques are model-specific or model-agnostic, and (iv) the explanations are conveyed through visualizations, surrogate models or taking into account features relevance.

The scientific literature reports an increasing amount of use cases where explainable artificial intelligence is applied. \cite{meister2021investigations} applied deep learning models to automate defect detection on composite components built with a fiber layup procedure. Furthermore, the authors explored using three Explainable Artificial Intelligence techniques (Smoothed Integrated Gradients (\cite{sundararajan2017axiomatic}), Guided Gradient Class Activation Mapping (\cite{shrikumar2017learning}) and DeepSHAP (\cite{selvaraju2017grad})), to understand whether the model has learned and thus can be trusted that it will behave robustly. \cite{senoner2021using} developed an approach to creating insights on how production parameters contribute to the process quality based on the estimated features' relevance to the forecast estimated with the Shapley additive explanations technique (\cite{lundberg2017unified}). Finally, \cite{serradilla2020interpreting} implemented multiple machine learning regression models to estimate the remaining life of industrial machinery and resorted to the Local Interpretable Model-Agnostic Explanations technique (\cite{ribeiro2016should}) to identify relevant predictor variables for individual and overall estimations. Given research had little consideration for the complexity of interactions between humans and their environment in manufacturing, much can be done to develop XAI approaches that consider anthropometrics, physiological and psychological states, and motivations to not only provide better explanations, but also enhance the workers' self-esteem and help them towards their self-actualization (\cite{lu2022outlook}). 

\subsubsection{Simulated reality}
Under simulated reality, we understand any program or process that can generate data resembling a particular aspect of reality. Such a process can take inputs and produce outputs, such as synthetic data or outcomes that reflect different scenarios or process changes. 

Machine learning models can solve complex tasks only if provided with data. Acquiring high-quality data can be a complex and expensive endeavor: lack of examples concerning faulty items for defect detection systems, wearing down and damaging a robotic system during data collection, or human errors when labeling the data are just some examples. Synthetic data is envisioned as a solution to such challenges. Much research is invested in making it easy to generate while avoiding annotation pitfalls, ethical and practical concerns and promising an unlimited supply of data (\cite{de2021next}). In addition, much research was invested in the past regarding synthetic data generation to cope with imbalanced datasets.

Nevertheless, the development of GANs opened a new research frontier, leading to promising results (\cite{creswell2018generative}). GANs consist of two networks: a generator (trained to map some noise input into a synthetic data sample) and a discriminator (that, given two examples, tries to distinguish the real from the synthetic one). This way, the generator learns to generate higher-quality samples based on the discriminator's feedback. While they were first applied to images (\cite{goodfellow2014generative}), models have been developed to enhance the quality of synthetic images and to apply them to other types of data, too (\cite{7796926,xu2019modeling}).

Simulated reality can be considered a key component of Reinforcement Learning. The reinforcement learning agent can explore an approximation of the real world through the simulator and learn efficient policies safely and without costly interactions with the world. Furthermore, by envisioning the consequences of an action, simulations can help to validate desired outcomes in a real-world setting (\cite{amodei2016concrete}).

Simulated reality has been applied in a wide range of manufacturing use cases. Neural Style Transfer (\cite{wei2020defective}) has been successfully used to generate synthetic samples by fusing defect snippets with images of non-defective manufactured pieces. Such images can be later used to enhance the algorithm's predictive capacity. Simulators have been widely applied to train Reinforcement Learning models in manufacturing. \cite{mahadevan1998optimizing} used them to simulate a production process and let the RL algorithm learn to maximize the throughput in assembly lines, regardless of the failures that can take place during the manufacturing process. \cite{oliff2020reinforcement} simulated human operators' performance under different circumstances (fatigue, shift, day of week) so that behavioral policies could be learned for robotic operators and ensure they provided an adequate response to the operators' performance variations. Finally, \cite{johannink2019residual} used Reinforcement Learning to learn robot control and evaluated their approach in both real-world and simulated environments. Bridging the gap from simulated to real-world knowledge remains a challenge.

\subsubsection{Intention Recognition in Manufacturing Lines}

The development of Industrial Internet of Things (IIoT) technologies and the availability of low-cost wearable sensors have enabled access to big data and the utilization of the sensors for the manufacturing industry (\cite{jeschke2017industrial}). Recently, various deep learning-based methods have been proposed to learn valuable information from big data and improve the effectiveness and safety of the manufacturing lines. In particular, worker's activity and intention recognition can be used for quantification and evaluation of the worker's performance and safety in the manufacturing lines. In addition, with the introduction of autonomous robots for effective manufacturing, efficient collaboration between robots and workers and the safety of workers are also becoming increasingly important.

Thanks to the miniaturization and reduction of costs, the adoption of wearable sensors has also been growing in the industrial context to investigate workers' conditions and well-being. Worker's health is a key factor in determining the organization's long-term competitiveness, and it is also directly related to production efficiency. The cumulative effect of positive impacts on the human factor brings economic benefit through productivity increase, scrap reduction, and decreased absenteeism. Few research works have been recently developed, where workers' physiological data are used to infer the insurgence of phenomena such as fatigue (\cite{maman2017data,maman2020data}) and mental stress (\cite{villani2020wearable}), which have a relevant impact on process performance. Another research line adopted eye-trackers, together with wearables and cameras, to estimate workers' attention and stress levels, understand assembly sequence, and identify the criticalities in the product design affecting the assembly process (\cite{peruzzini2017benchmarking}).

Human-robot collaboration in open workspaces is realized through cobots, for which additional mechanisms must be developed to ensure the workers' safety, given that humans can be easily hurt in case of contact due to the large workload or moving mass (\cite{bi2021safety}). To realize such collaboration, human movement prediction is of utmost importance to avoid collisions and minimize injuries caused by such collisions (\cite{buerkle2021eeg}). Many researchers have developed various activity and intention recognition methods by using machine learning-based algorithms and wearable sensors. \cite{malaise2018activity} proposed activity recognition with Hidden Markov Model (HMM)-based models and multiple wearable sensors for a manufacturing scenario. \cite{tao2018worker} proposed an activity recognition method using Inertial Measurement Unit (IMU) and surface electromyography (sEMG) signals obtained from a Myo armband. They combined the IMU and sEMG signals and fed them into convolutional neural networks (CNN) for worker activity classification. \cite{kang2018motion} developed a motion recognition system for worker safety in manufacturing work cells, leveraging a vision system.  \cite{forkan2019industrial} introduced an IIoT solution for monitoring, evaluating, and improving worker and related plant productivity based on worker activity recognition using a distributed platform and wearable sensors. \cite{gunther2019activity} proposed a human activity recognition approach to detect assembly processes in a production environment by tracking activities performed with tools. \cite{tao2020multi} proposed a multi-modal activity recognition method by leveraging information from different wearable sensors and visual cameras. Finally, \cite{buerkle2021eeg} proposed using a mobile electroencephalogram and machine learning models to forecast operators' movements based on two neurophysiological phenomena that can be measured before the actual movement takes place: (a) a weak signal that occurs about 1.5s before a movement, and (b) a strong signal that occurs about 0.5s before any movement.

\subsubsection{Conversational interfaces}
Spoken dialog systems and conversational multimodal interfaces leverage artificial intelligence and can reduce friction and enhance human-machine interactions (\cite{klopfenstein2017rise,vajpai2016industrial,maurtua2017natural}) by approximating a human conversation. However, in practice, conversational interfaces mostly act as the first level of support and cannot offer much help as a knowledgeable human. They can be classified into three broad categories: (i) basic-bots, (ii) text-based assistants, and (iii) voice-based assistants. While basic bots have a simple design and allow basic commands, the text-based assistants (also known as chatbots) can interpret users' text and enable more complex interactions. Both cases require speech-to-text and text-to-speech technologies, especially if verbal interaction with the conversational interface is supported. Many tools have been developed to support the aforementioned functionalities. Among them, we find the Web Speech API\footnote{\url{https://developer.mozilla.org/en-US/docs/Web/API/Web_Speech_API}}, which can be configured to recognize expressions based on a finite set of options defined through a grammar \footnote{\url{https://www.w3.org/TR/jsgf/}}. Most advanced version of conversational interfaces are represented by voice assistants, such as the Google Assistant\footnote{\url{https://assistant.google.com/}}, Apple's Siri\footnote{\url{https://www.apple.com/siri/}}, Microsoft's Cortana\footnote{\url{https://www.microsoft.com/en-in/windows/cortana}}, or Amazon's Alexa\footnote{\url{https://developer.amazon.com/alexa}}. They can be integrated into multiple devices and environments through publicly available application development interfaces (APIs), enabling new business opportunities (\cite{8430412}). Given voice interfaces can place unnecessary constraints in some use cases, they can be complemented following a multimodal approach (\cite{DBLP:conf/aaate/Kouroupetroglou17}).

A few implementations were described in an industrial setting. \cite{silaghi2014voice} researched the use of voice commands in noisy industrial environments, showing that noises can be attenuated with adequate noise filtering techniques. \cite{wellsandta2020concept} developed an intelligent digital assistant that connects multiple information systems to support maintenance staff on their tasks regarding operative maintenance. They exploit the fact that access to the voice assistant's functions is hand free and that voice operation is usually faster than writing. \cite{afanasev2019concept} developed a method to integrate a voice assistant and modular cyber-physical production system, where the operator could request help to find out-of-sight equipment or get specific sensor readings. Finally, \cite{li2022bringing} developed a virtual assistant to assist workers on dangerous and challenging manufacturing tasks, controlling industrial mobile manipulators that combine robotic arms with mobile platforms used on shop floors. The assistant uses a language service to extract keywords, recognize intent, and ground knowledge based on a knowledge graph. Furthermore, conversation strategies and response templates are used to ensure the assistant can respond in different ways, event when the same question is asked repeatedly.

\subsubsection{Security}
While the next generation of manufacturing aims to incorporate a wide variety of technologies to enable more efficient manufacturing and product lifecycles, at the same time, the attack surface increases, and new threats against confidentiality, integrity, and availability are introduced (\cite{chhetri2017security,chhetri2018manufacturing}). These are exacerbated by the existence of a large number of legacy equipment, the lack of patching and continuous updates on the industrial equipment and infrastructure, and the fact that cyberattacks on cyber-physical systems achieve a physical dimension, which can affect human safety (\cite{elhabashy2019cyber}). Artificial intelligence has proved its efficiency for threat intelligence sensing, intrusion detection, and malware classification, while how to ensure a model itself was not compromised remains a topic of major research (\cite{conti2018cyber,li2018cyber}).

Multiple cyberattack case studies in manufacturing have been analyzed in the scientific literature. \cite{zeltmann2016manufacturing} studied how embedded defects during additive manufacturing can compromise the quality of products without being detected during the quality inspection procedure. The attacks can be fulfilled by either compromising the CAD files or the G-codes. \cite{ranabhat2019optimal} demonstrated sabotage attacks on carbon fiber reinforced polymer by identifying critical force bearing plies and rotating them. Therefore, the resulting compromised design file provides a product specification that renders the manufactured product useless. Finally, \cite{liu2020poisoning} describe a data poisoning attack through which the resulting machine learning model is not able to detect hotspots in integrated circuit boards.

In order to mitigate the threats mentioned above, steps must be taken to prevent the attacks, detect their effects, and respond, neutralizing them and mitigating their consequences (\cite{elhabashy2019cyber}). On the prevention side, \cite{wegner2017new} advocated for the extensive use of authentication and authorization in the manufacturing setting. To that end, the authors proposed using asymmetric encryption keys to enable encrypted communications, a \textit{comptroller} (software authorizing actions in the manufacturing network, and encryption key provider) to ensure the input data is encoded and handled to a Manufacturing Security Enforcement Device, which then ensures the integrity of the transmitted data. A security framework for cyber-physical systems was proposed by \cite{wu2018dacdi}, defining five steps: Define, Audit, Correlate, Disclose, Improve (DACDI). \textit{Define} refers to the scope of work, considering the architecture, the attack surface, vector, impact, target, and consequence, and the audit material. \textit{Audit} relates to the process of collecting cyber and physical data required for intrusion detection. Artificial intelligence is being increasingly used in this regard, leveraging paradigms such as active learning to combine machine and human strengths (\cite{klein2022jasmine}). \textit{Correlate} attempts to establish relationships between cyber and physical data considering time and production sequences, the scale and duration of the attack, and therefore reduce the number of false positives and assist in identifying the root causes of alerts. \textit{Disclose} establishes a set of methods used to stop the intrusion as quickly as possible. Finally, \textit{Improve} aims to incrementally enhance the security policies to avoid similar issues in the future. Another approach was proposed by \cite{bayanifar2017enhancing}, who described an agent-based system capable of real-time supervision, control, and autonomous decision-making to defend against or mitigate measured risks.

\section{Safe, Trusted, and Human-Centered Architecture}\label{S:REFERENCE-ARCHITECTURE}

\subsection{Architecture Values-Based Principles}

\begin{figure*}[!t]
\centering
\includegraphics[width=0.6\textwidth]{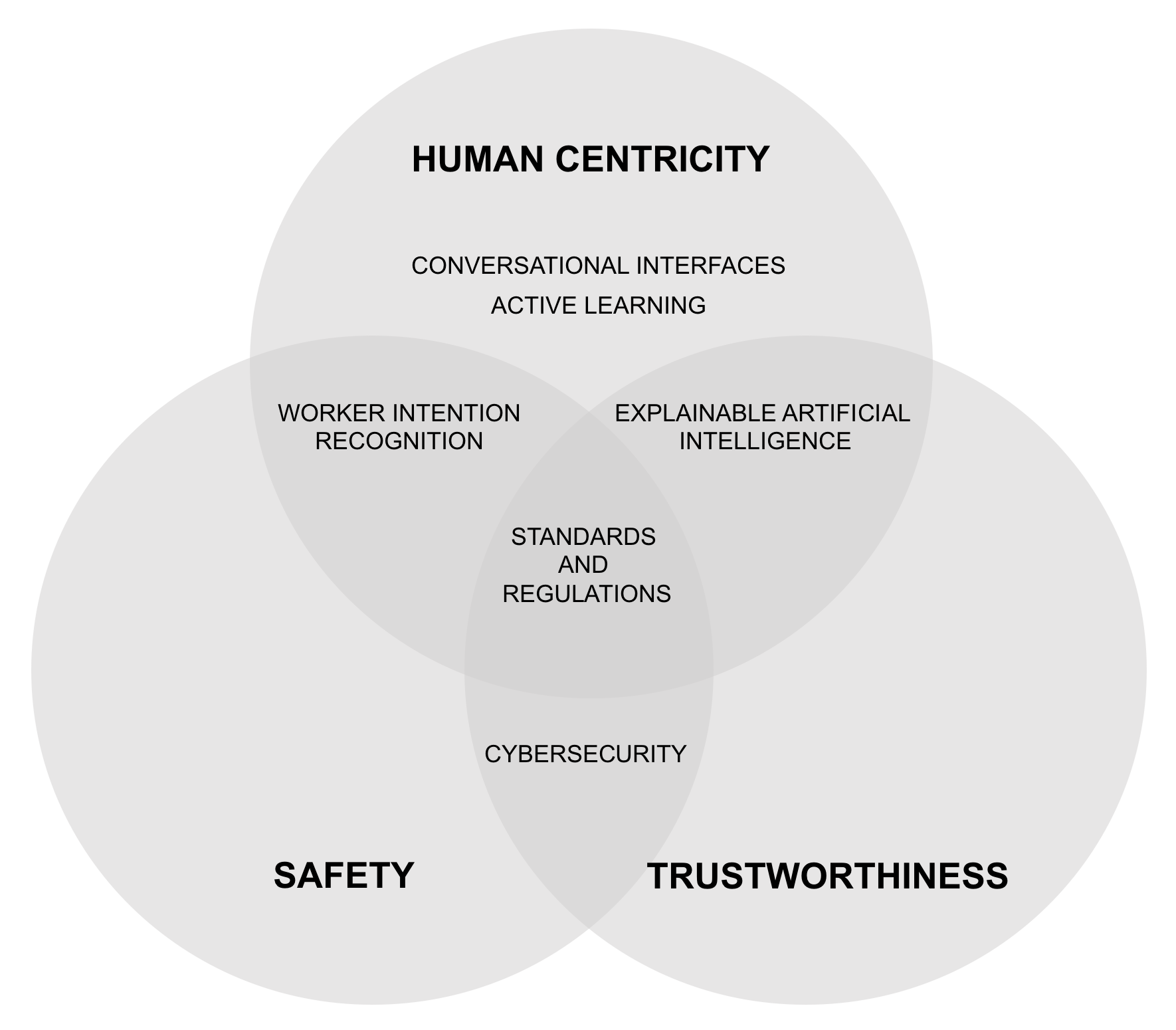}
\caption{
Caption: Intersection of architecture value-based principles, and architecture building blocks addressing them.
Alt Text: Three intersected circles describing the value-based principles for Industry 5.0 and architecture aspects realized in their intersection. 
}
\label{F:REFERENCE-ARCHITECTURE-VALUE-BASED-PRINCIPLES}
\end{figure*}

The proposed architecture is designed to comply with three key desired characteristics for the manufacturing environments in Industry 5.0: safety, trustworthiness, and human centricity. \textit{Safety} is defined as the condition of being protected from danger, risk, or injury. In a manufacturing setting, \textit{safety} can refer either to \textit{product safety} (quality of a product and its utilization without risk), or \textit{human safety} (accident prevention in work situations), and the injuries usually relate to occupational accidents, or bad ergonomics (\cite{wilson2005impact,sadeghi2016design}). Trustworthiness is understood as the quality of deserving trust. In the context of manufacturing systems, it can be defined as a composite of transparency, reliability, availability, safety, and integrity (\cite{yu2017trustworthiness}). In manufacturing, trustworthiness refers to the ability of a manufacturing system to perform as expected, even in the face of anomalous events (e.g., cyberattacks), and whose inner workings are intelligible to the human persons who interact with them. 
Human-centricity in production systems refers to designs that put the human person at the center of the production process, taking into account their competencies, needs, and desires, and expecting them to be in control of the work process while ensuring a healthy and interactive working environment (\cite{may2015new}). We consider \textit{Safety} and \textit{Trustworthiness} are critical to a human-centric approach, and therefore render them as supporting pillars of the \textit{Human Centricity} values-based principle in Fig. \ref{F:REFERENCE-ARCHITECTURE-VALUE-BASED-PRINCIPLES}.

We depict the above-listed architecture value-based principles in Fig. \ref{F:REFERENCE-ARCHITECTURE-VALUE-BASED-PRINCIPLES}, and how do the building blocks, detailed in Section \ref{S:RELATED-WORK}, relate to them. \textit{Cybersecurity} is considered at the intersection of safety and trustworthiness since it ensures manufacturing systems and data are not disrupted through cybersecurity attacks (e.g., data poisoning or malware attacks). The \textit{Worker Intention Recognition} is found at the intersection of safety and human centricity since it aims to track better and understand the human person to predict its intentions (e.g., movements) and adapt to the environment according to this information. \textit{Explainable Artificial Intelligence} provides insights regarding the inner workings of artificial intelligence models and therefore contributes to the trustworthiness while being eminently human-centric. \textit{Conversational Interfaces} and \textit{Active Learning} place the human person at their center, either by easing interactions between humans and machines or seeking synergies between their strengths to enhance Artificial Intelligence models' learning. Finally, \textit{Standards and Regulations} are considered at the intersection of the three aforementioned value-based principles, given they organize and regulate aspects related to each of them.

\subsection{Architecture for Safe, Trusted, and Human-Centric Manufacturing Systems}

\begin{figure*}[!t]
\centering
\includegraphics[width=\textwidth]{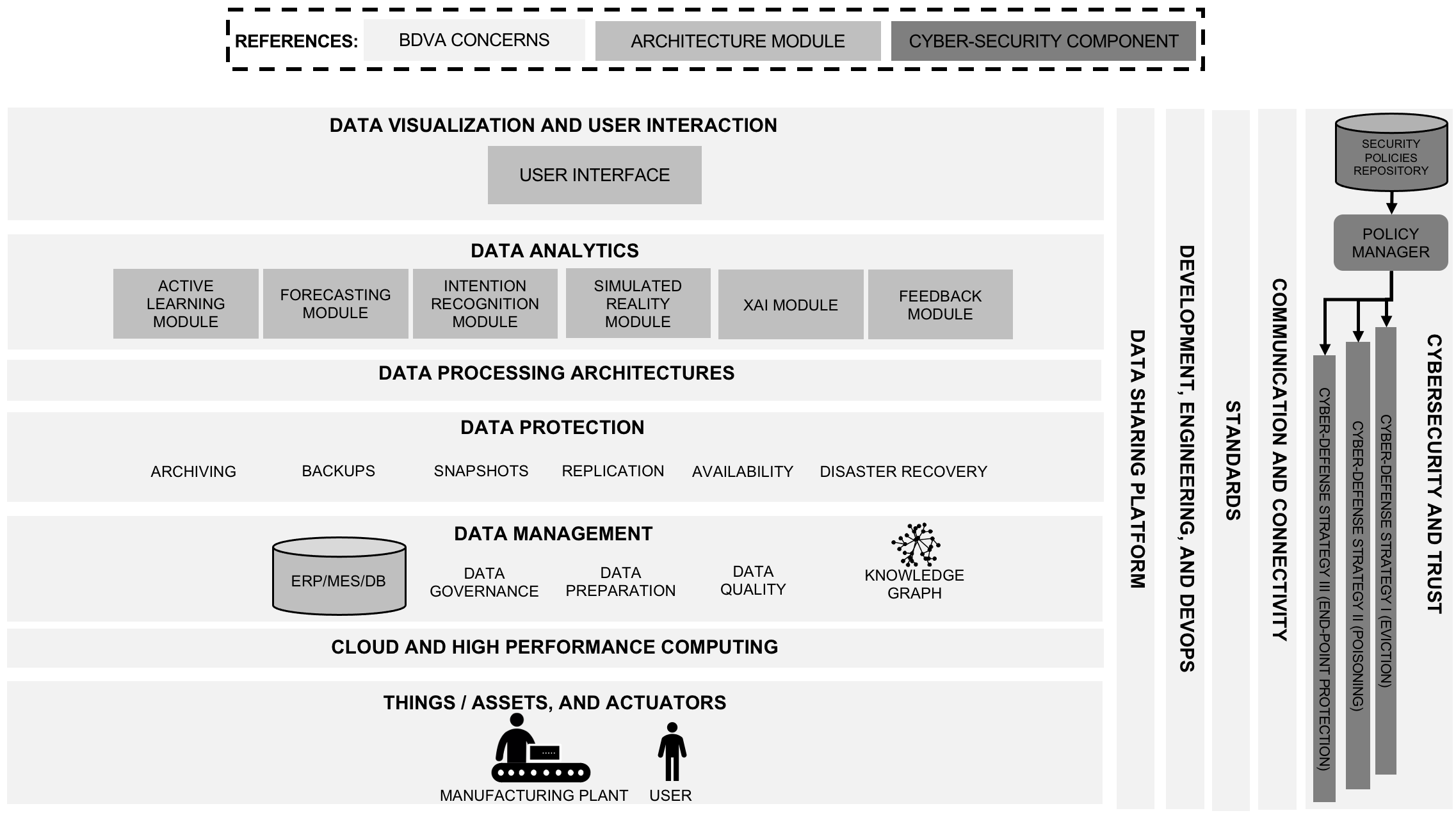}
\caption{
Caption: Proposed architecture contextualized within the BDVA reference architecture components.
Alt Text: Architecture diagram showing architecture modules and BDVA concerns.}
\label{F:REFERENCE-ARCHITECTURE-BDVA}
\end{figure*} 

We propose a modular architecture for manufacturing systems, considering three core value-based principles: safety, trustworthiness, and human centricity. The proposed architecture complies with the BDVA reference architecture (see Fig. \ref{F:REFERENCE-ARCHITECTURE-BDVA}) and considers cybersecurity a transversal concern, which can be implemented following guidelines from the IISF or ISO 27000, along with other security frameworks and standards. The cybersecurity layer transversal implements a \textit{Security Policies Repository} and a \textit{Policy Manager}. The \textit{Security Policies Repository} associates risk-mitigation and cyber defense strategies to potential vulnerabilities and specific cyberattacks. The \textit{Policy Manager}, on the other hand, configures security policies and ensures they are deployed, changing the security operations.

\begin{figure*}[!t]
\centering
\includegraphics[width=\textwidth]{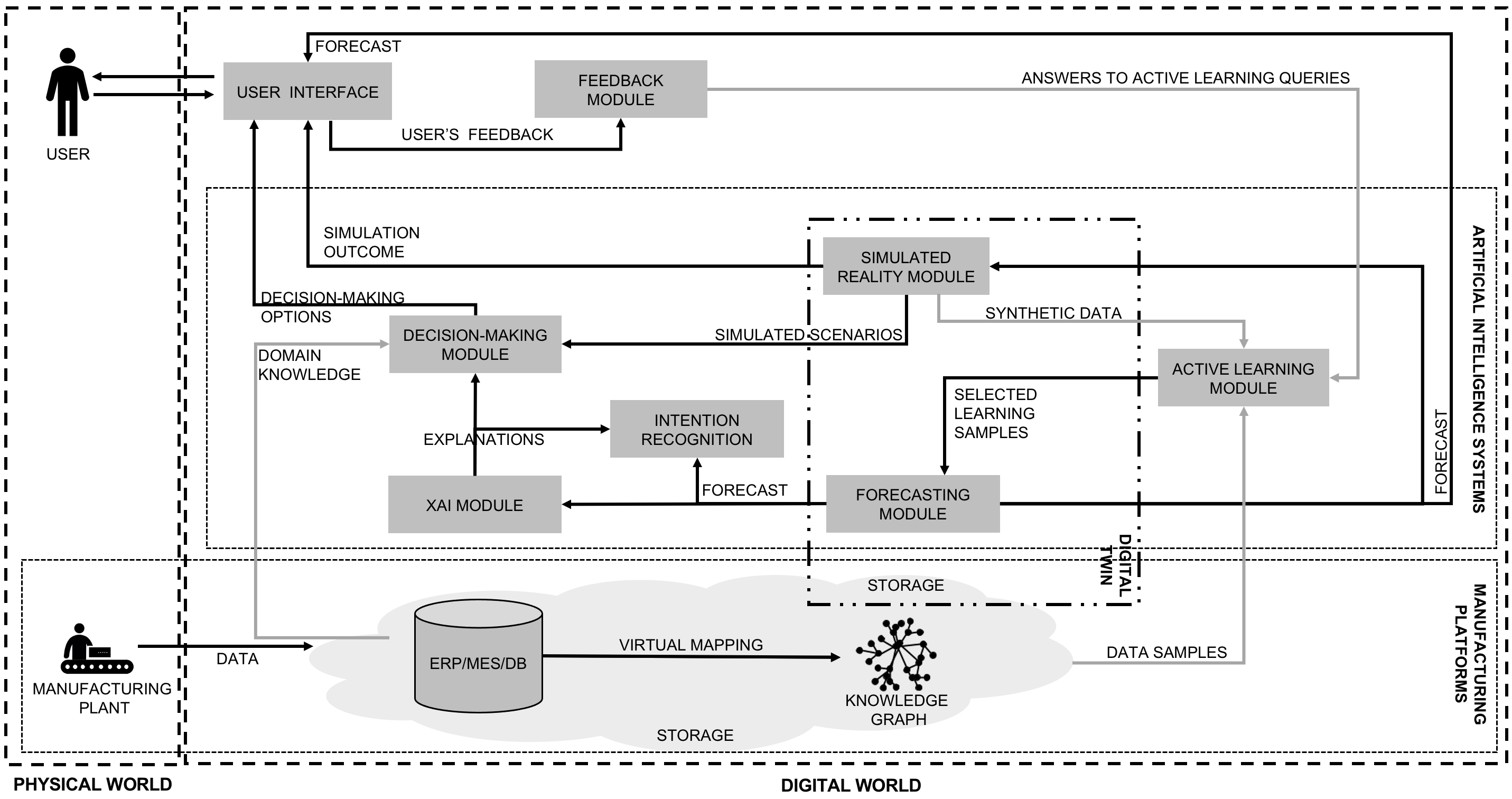}
\caption{
Caption: The proposed architecture modules, a storage layer, and their interactions. In addition, we distinguish (a) the physical and digital worlds, (b) manufacturing platforms, (c) artificial intelligence systems, and (d) digital twin capabilities.
Alt Text: Architecture diagram showing the interaction between the proposed architecture modules.
}
\label{F:REFERENCE-ARCHITECTURE-LOGICAL}
\end{figure*}

The architecture evolved organically from a set of use cases developed for two EU H2020 projects. It comprises the following modules, whose interaction is depicted in Fig. \ref{F:REFERENCE-ARCHITECTURE-LOGICAL}:

\begin{itemize}
    \item \textbf{Simulated Reality Module}: uses heuristics, statistical, and machine learning models to either create alternative scenarios or generate synthetic data. Synthetic data is frequently used to mitigate the lack of data, either by replacing expensive data gathering procedures or enriching the existing datasets. On the other hand, the simulated scenarios are frequently used in Reinforcement Learning problems to foster models' learning while avoiding the complexities of a real-world environment. Furthermore, simulations can also be used to project possible outcomes based on potential users' decisions. Such capability enables what-if scenarios, which can be used to inform better decision-making processes. The \textit{Simulated Reality Module} provides synthetic data instances to the \textit{Active Learning Module}, simulated scenarios to the \textit{Decision-Making Module}, and simulation outcomes to the user (through a \textit{User Interface}).
    \item \textbf{Forecasting Module}: provides forecasts for a wide range of manufacturing scenarios, leveraging artificial intelligence and statistical simulation models. The outcomes of such models depend on the goal to be solved (e.g., classification, regression, clustering, or ranking). While machine learning models require data to learn patterns and create inductive predictions, simulation models can predict future outcomes based on particular heuristic and configuration parameters that define the problem at hand. The \textit{Forecasting Module} can receive inputs either from the storage or the \textit{Active Learning Module}. At the same time, it provides forecasts to the user, the \textit{Simulated Reality Module}, and the \textit{XAI Module}. With the former, it can also share relevant information regarding the forecasting model to facilitate the creation of accurate explanations.
    \item \textbf{XAI Module}: is concerned with providing adequate explanations regarding artificial intelligence models and their forecasts. Such explanations aim to inform the user regarding the models' rationale behind a particular forecast and must be tailored to the users' profile to ensure the appropriate vocabulary, level of detail, and explanation type (e.g., feature ranking, counterfactual explanation, or contrastive explanation) is provided. Furthermore, the module must ensure that no sensitive information is exposed to users who must not have access to it. Finally, the explanations can be enriched with domain knowledge and information from complementary sources. Such enrichment can provide context to enhance users' understanding and, therefore, enable the user to evaluate the forecast and decision-making. The \textit{XAI Module} provides input to the \textit{Decision-Making Module} and explanations to the user.
    \item \textbf{Decision-Making Module}: is concerned with recommending decision-making options to the users. Envisioned as a recommender system, it can leverage expert knowledge and predictions obtained from inductive models and simulations and exploits it using heuristics and machine learning approaches. Given a particular context, it provides the user with the best possible decision-making options available to achieve the desired outcome. It receives input from the \textit{XAI Module} and \textit{Forecasting Module} and can retrieve expert knowledge encoded in the storage (e.g., a knowledge graph). 
    \item \textbf{Active Learning Module}: implements a set of strategies to take actions on how data must be gathered to realize a learning objective. In supervised machine learning models, this is realized by selecting unlabeled data, which can lead to the best models' learning outcomes, and request labels to a human annotator. Another use case can be the data gathering that concerns a knowledge base enrichment. To that end, heuristics can be applied to detect missing facts and relationships ask for and store locally observed collective knowledge not captured by other means (\cite{preece2015sherlock}). The \textit{Active Learning Module} interacts with the storage and the \textit{Simulated Reality Module} to retrieve data, and the \textit{Feedback Module} to collect answers to queries presented to the user.
    \item \textbf{Feedback Module}: collects feedback from users, which can be either explicit (a rating or an opinion) or implicit (the lack of feedback can be itself considered a signal) (\cite{oard1998implicit}). The feedback can refer to feedback regarding given predictions from the \textit{Forecasting Module}, explanations provided by the \textit{XAI Module}, or decision-making options recommended by the \textit{Decision-Making Module}. It directly interacts with the \textit{Active Learning Module} and the user (through the \textit{User Interface}), and indirectly (registering and storing the feedback) with other modules' feedback functionality exposed to the user.
    \item \textbf{Intention Recognition Module}: predicts the user's movement trajectory based on artificial intelligence models and helps to decide whether the mobile robot should move faster, slower, or completely stop. The module receives sensor and video data. The data is captured by cameras in the manufacturing line or by sensors attached to the user's body. The recognized worker's activity and intention can be used by the \textit{Forecasting Module} and the \textit{XAI Module} to decide the following action of the mobile robot in the manufacturing line.
    \item \textbf{User interface}: enables users' multimodal interactions with the system, e.g., complementing the voice interactions with on-screen forms. Furthermore, it enables the machine to provide information to the user through audio, natural language, or other means such as visual information.
\end{itemize}

Interactions with the human persons are realized through a \textit{User Interface}, while the data is stored in a \textit{Storage}, which can be realized by different means (e.g., databases, filesystem, knowledge graph) based on the requirements of each module. The \textit{User Interface} can be implemented taking into account multiple modalities. While the use of graphical user interfaces is most extended, there is increasing adoption of voice agents.

Based on the modules described above, multiple functionalities can be realized. While the \textit{Storage} can store near data collected from the physical world, the \textit{Simulated Reality Module} and the \textit{Forecasting Module} can provide behavior to Digital Twins mirroring humans (e.g., to monitor fatigue or emotional status), machines (e.g., for predictive maintenance), and manufacturing processes (e.g., supply chain and production). Furthermore, the \textit{Forecasting Module} can be used to recognize and predict the workers' intention and expected movement trajectories. This information can then be used to adapt to the environment, e.g., by deciding whether autonomous mobile robots should move faster, slower, or completely stop. Finally, the \textit{Active Learning Module} and \textit{Explainable Artificial Intelligence Module} can be combined to create synergic relationships between humans and machines. While the \textit{Active Learning Module} requires the human to provide expert knowledge to the machines and teach them, the \textit{Explainable Artificial Intelligence Module} enables the humans to learn from machines.

\section{Validating use cases}\label{S:USE-CASES-OVERVIEW}
We propose three validating use cases (demand forecasting, quality inspection, and intention recognition), which align with the human-centricity, trustworthiness, and safety Industry 5.0 core values. Through the first two use cases, we aimed to realize the research gap highlighted by \cite{lu2022outlook}, considering the interactions between human beings and their environment, developing a collaborative dynamic between humans and machines leveraging active learning and explainable artificial intelligence along with other relevant technologies. With the third use case, we aim to describe how intention recognition is considered in the proposed architecture to enable safe collaboration with cobots (\cite{hentout2019human}). Finally, we evaluated the proposed architecture through the proposed use cases, assessing the internal validity (the results obtained can only be attributed to the manipulated variable) and external validity (which describes the generalizability of our architecture, demonstrated through three use cases) (\cite{salkind2006exploring}). It must be noted, that the components were implemented separately and not deployed into a productive environment.

\subsection{Demand Forecasting}\label{SS:USE-CASES-DEMAND-FORECASTING}
Research regarding demand forecasting was performed with data provided by a European original equipment manufacturer targeting the global automotive industry market. 
Demand forecasting aims to estimate future customer demand over a defined period of time, using historical data and other sources of information.
The ability to accurately forecast future demand allows to reduce operational inefficiencies (e.g., high stocks or stock shortages), which have a direct impact on goods produced in the supply chain and therefore can affect the brands' reputation (\cite{bruhl2009sales}). Furthermore, insights into future demand enable better decision-making on various levels (e.g., regarding resources, workers, manufactured products, and logistics) (\cite{thomopoulos2015demand}).
While human forecasts are prone to multiple biases, the statistical and machine learning models can be used to learn patterns present in data obtained from multiple information sources to create accurate forecasts (\cite{corredor2014cognitive,hogarth1981forecasting}). Such models do not replace humans but provide a means to establish a synergic relationship. The model provides a forecast, and the user can make judgmental adjustments and make decisions based on them. Judgemental adjustments need to be done when some information is not available to the model, e.g., knowledge regarding future and extraordinary events that the model cannot capture from the existing information sources (\cite{fildes2021stability}). Furthermore, it is recommended to record such forecast adjustments and the reasons behind them to be evaluated in retrospect. Such records can provide valuable input to improve the demand forecasting models. It must be noted that the demand forecasting models, regardless of their accuracy, must be considered tools to ease the planning duties. The planners are responsible for the decisions taken, regardless of the forecast outcomes.

Given that the planners hold responsibility for their decisions, the forecast must be complemented with insights regarding the models' rationale to enable responsible decision-making (\cite{almada2019human,wang2020toward}). Furthermore, an explanation regarding a particular forecast is sometimes legally required (\cite{goodman2017european}). Such insights can be either derived from the model or attained through specific techniques. The insights can be served as explanations to the users, tailoring them according to their purpose and the target stakeholders (\cite{samek2019towards}). Such explanations must convey enough and relevant information, resemble a logical explanation (\cite{doran2017does,pedreschi2018open}), focus on aspects the stakeholder can act on to change an outcome (\cite{verma2020counterfactual,keane2021if}), and ensure confidentiality is preserved (\cite{rovzanec2022knowledge}). Good decision-making can require not only understanding the models' rationale but having relevant domain knowledge and contextual information at hand too (\cite{arrieta2020explainable,rozanec2021explainable,zajec2021help}). Furthermore, the users' perception of such explanations must be assessed to ensure their purpose is achieved (\cite{mohseni2021multidisciplinary,sovrano2021objective}).

\subsection{Quality Inspection}\label{SS:USE-CASES-QUALITY-INSPECTION}

\begin{figure*}[!t]
\centering
\includegraphics[width=0.8\textwidth]{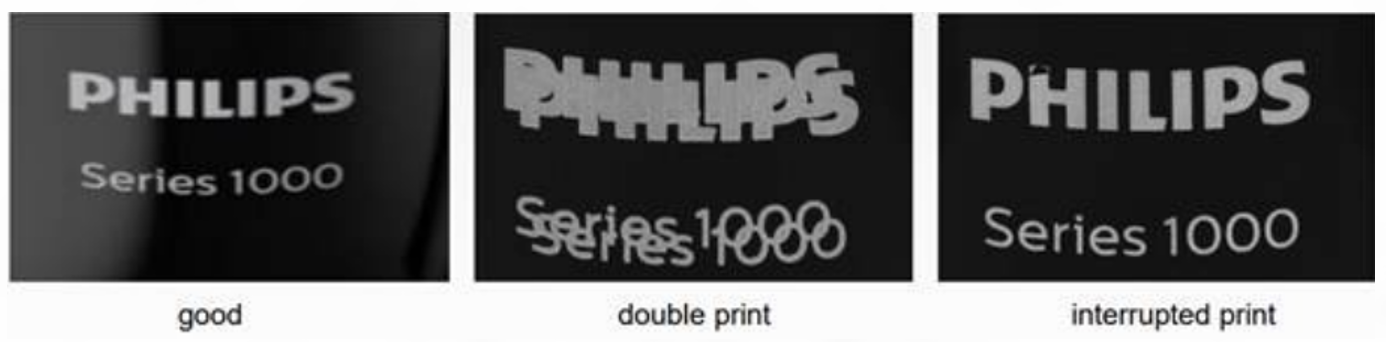}
\caption{
Caption: Samples of three types of images: (a) good (no defect), (b) double-print (defect), and (c) interrupted print (defect).
Alt Text: Sample images regarding defective and non-defective manufactured parts for the Philips use case.
}
\label{F:PHIA-DEFECTS}
\end{figure*}

Research regarding quality inspection was performed with data provided by \textit{Philips Consumer Lifestyle BV}. The dataset consisted of images focused on the company's logo printed on manufactured shavers. 
The visual quality inspection aims to detect defective logo printing on the shavers, focusing on printing pads used for a wide range of products and logos. Currently, two types of defects are classified related to the printing quality of the logo on the shaver: double printing and interrupted printing. Handling, inspecting and labeling the products can be addressed with robotics and artificial intelligence. It is estimated that automating the process mentioned above could speed it up by more than 40\%, while the labeling effort can be alleviated by incorporating active learning  (\cite{trajkova2021active,rovzanec2021streaming}).

When addressing the quality inspection use case, many challenges must be solved. We focused on four of them: (i) automate the visual inspection, (ii) address data imbalance, (iii) understand the models' rationale behind the forecast, and (iv) enhance the manual revision process. While class imbalance is natural to quality inspection problems, its acuteness only increases over time: the greater the quality of the manufacturing process, the higher the scarcity of defective samples will be. The class imbalance has at least two implications. First, the increasing scarcity of defective parts affects the amount of data available to train defect detection models, affecting the capacity to improve them. Second, the higher the imbalance between good and defective products, the higher the risk that the inspection operators will not detect defective parts due to fatigue. It is thus necessary to devise mechanisms to mitigate such scenarios, ensuring high-quality standards are met while also enhancing the operators' work experience.

\subsection{Human Behaviour Prediction and Safe Zone Detection}
SmartFactoryKL is an industry test-bed owned by the Deutsches Forschungszentrum f\"ur K\"unstliche Intelligenz (DFKI), which demonstrates the latest technologies from the industry domain building industry-standard demonstrators. One of the demanding aspects to our demonstrators would be adding safety considerations while working with an autonomous robot while combining artificial intelligence technologies. In this context, SmartFactoryKL aims to improve its demonstrator to have a high production rate while keeping workers and the hardware equipment safe using AI technologies. 

In order to achieve this, three use cases were initialized for this test-bed:
\begin{enumerate}
\item Human intention recognition
\item Robot reconfiguration based on the dynamic layout
\item Dynamic path planning using both aforementioned use cases. 
\end{enumerate}

The first use case aimed to detect human activities and predict their next actions, which then will be combined with robot navigation to create a safer environment. For this matter, DFKI initialized some workflows as the typical worker scenarios happening during regular daily work. For the training phase, the behaviors of more than ten participants were recorded, who were supposed to follow the same or similar flows. The recordings were made using wrist sensors which are then analyzed in detail to detect different activities they are currently performing. The use case converts algorithm-based human activities into an artificial intelligence approach and predicts their following actions based on their daily activities.

The second use case aims to dynamically update the navigation route of the mobile robot by considering static and dynamic objects in the environment, which can be human and other (non-)moving objects. A desired outcome of the use case is the ability to ease the robot reconfiguration given the environment layout (including the production stations) changes.

Finally, both use cases mentioned above were combined into a third one to ensure a safe environment for workers and hardware equipment. The newly received coordinates of the stations were used to set the robot's destinations. The speed of the robot and the objects in the layout were also considered to create a collision-free navigation path for the robot. Moreover, predicting human behavior is essential to configure the mobile robot, avoid possible collisions, and create safety zones. Human movement intention forecasting was used to plan an optimal robot route without collision risks. Such a route relies on the defined workflow progress done by workers to respond to production requirements. The use case's goal is to keep the production level high and maintain human safety during robot activity.

\section{Experiments and Results}\label{S:EXPERIMENTS-AND-RESULTS}

\subsection{Demand Forecasting}
We addressed the demand forecasting use case in four parts: (i) development of forecasting models, (ii) models' explainability, (iii) decision-making options recommendation, and (iv) the development of a voice interface. 

To enable demand forecasts, we developed multiple statistical and machine learning models for products with smooth and erratic demands (\cite{rovzanec2021automotive}) and products with lumpy and intermittent demand (\cite{rovzanec2021reframing}) (\textit{Forecasting Module}). The models were developed based on real-world data from a European original equipment manufacturer targeting the global automotive industry market. For products with smooth and erratic demand, we found that the best results were obtained with global models trained across multiple time series, assuming that there is enough similarity between them to enhance learning. Furthermore, our research shows that the forecast errors of such models can be constrained by pooling product demand time series based on the past demand magnitude. On the other hand, for products with lumpy and intermittent demand we found best results were obtained applying a two-fold approach, which was more than 30\% more precise than best existing approaches when predicting demand occurrence, resulting in important gains when considering Stock-keeping-oriented Prediction Error Costs (\cite{martin2020new}).

Demand forecasts influence the supply chain managers' decision-making process, and therefore additional insights, obtained through Explainable Artificial Intelligence (\textit{XAI Module}), must be provided to understand the model's rationale behind a forecast. To that end, we explored the use of surrogate models to understand which features were most relevant to a particular forecast and used a custom ontology model to map relevant concepts to the aforementioned features (\cite{rozanec2021explainable,rovzanec2022knowledge,rovzanec2022enriching}). Such mapping hides sensitive details regarding the underlying model from the end-user. It ensures meaning is conveyed with high-level concepts intelligible to the users while remaining faithful to the ranking of the features. Furthermore, we enriched the explanations by providing media news information regarding events that could have influenced the demand in the past and searched for open datasets that could be used to enrich the models' data to lead to better results in the future. Our demand forecasting models achieved state-of-the-art performance, while the enriched explanations displayed a high-degree of precision: for the worst cases, we achieved a precision of 0,95 for the media events displayed, a precision of 0,71 for the media keywords, and a precision of 0,56 for datasets displayed to the users.

Finally, we developed a heuristic recommender system to advise logisticians on decision-making options based on the demand forecasting outcomes (\cite{rovzanec2021xai}) (\textit{Decision-making Module}). The prototype application supported gathering (i) feedback regarding existing decision-making options (\textit{Feedback Module}) and (ii) new knowledge to mitigate scenarios where the provided decision-making options did not satisfy the user. Feedback and new knowledge were persisted into a knowledge graph modeled after an ontology developed for this purpose. Furthermore, the user interface was developed to support interactions either through a graphical user interface or voice commands\footnote{A video of the application was published in \url{https://www.youtube.com/watch?v=EpFBNwz6Klk}}. Future work will develop voice interfaces that are robust to noisy industrial environments.

Among the main challenges faced to create the demand forecasting models and provide models' explainability were the data acquisition and ensuring data quality. Data acquisition required working with different environments, application programming interfaces (e.g., to retrieve media news, information regarding demand, or other complementary information), and query constraints related to those environments and interfaces. Multiple iterations were performed to validate successive model versions and information displayed to the experts. Furthermore, through the iterations we worked on enhancing the models' performance, and expand the models' scope towards a greater number of products.

\subsection{Quality Inspection}

\begin{figure*}[!t]
\centering
\includegraphics[width=0.8\textwidth]{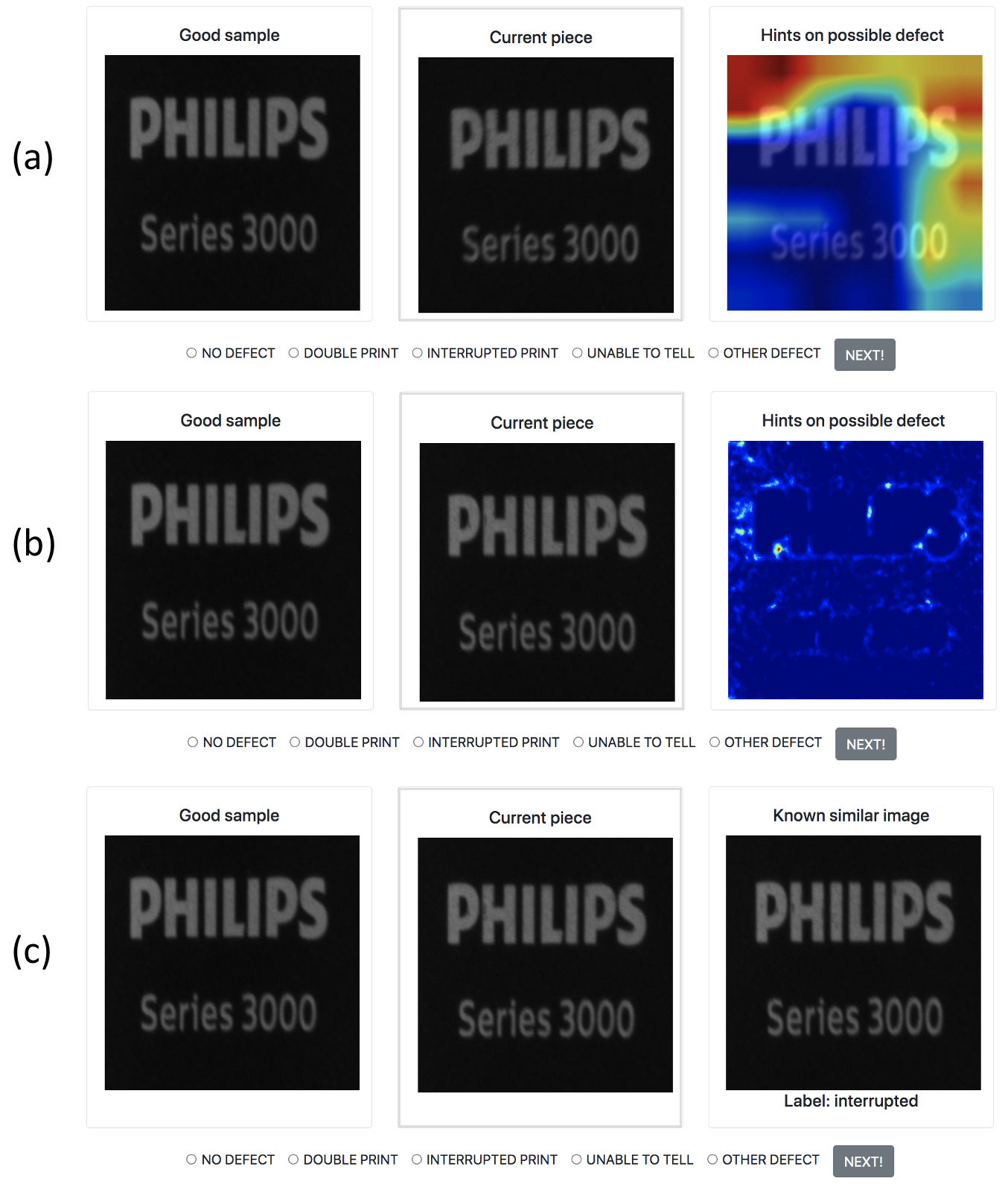}
\caption{
Caption: Sample screen for the manual revision process. We provide the operator an image of a non-defective part, the image of the component being inspected, and the hints regarding where we do expect the error can be. The images correspond to cases where the hints were created with (a) GradCAM, (b) DRAEM, and (c) the most similar labeled image.
Alt Text: Images showing different defect hinting approaches.
}
\label{F:PHIA-HINTS}
\end{figure*}

Our work regarding quality inspection was addressed in four parts: (i) development of machine learning models for automated visual inspection of manufactured products (\textit{Forecasting Module}), (ii) use of active learning to reduce the manual labeling efforts (\textit{Active Learning Module}), (iii) use of simulated reality to generate synthetic images (\textit{Simulated Reality Module}), and (iv) explore techniques to hint the user where defect could be (some hints were retrieved from the \textit{XAI Module}).

To automate visual inspection, we explored batch and streaming models (\cite{trajkova2021active,rovzanec2021streaming}). While the batch models usually achieve better performance, they cannot leverage new data as it becomes available, but rather a new retrained model has to be deployed. Furthermore, while using all available data can help the model achieve better discriminative power, it is desirable to minimize the labeling and manual revision efforts, which can be achieved through active learning. We found that while models trained through active learning had a slightly inferior discriminative power, their performance consistently improved over time. In an active learning setting, we found that the best batch model (multilayer perceptron) achieved an average performance of 0,9792 AUC ROC, while the best streaming model (streaming kNN) lagged by at least 0,16 points. Both models were built using a ResNet-18 model (\cite{he2016deep}) to extract embeddings from the Average Pooling layer. We selected a subset of features based on their mutual information ranking and evaluated the models with a stratified 10-fold cross-validation (\cite{zeng2000distribution}). Given the performance gap between both types of models and the high cost of miss-classification, batch models were considered the best choice in this use case. To decouple specific model implementations and their predictions from any service using those predictions, machine learning models can be calibrated to produce calibrated probabilities. We noted that in some cases, such model calibration further enhanced the models' discriminative power (\cite{rovzanec2022active}).

Given that defective parts always concern a small proportion of the overall production, it would be natural that the datasets are skewed, having a strong class imbalance. Furthermore, such imbalance is expected to increase over time as the manufacturing quality improves. Therefore, the \textit{Simulated Reality Module} was used to generate synthetic images with two purposes. First, they were used to achieve greater class balance, leading to nearly perfect classification results (\cite{rozanecqa2:inreview}). Second, synthetic images were used to balance data streams in manual revision to ensure attention is maximized and that defective pieces are not dismissed as good ones due to inertia (\cite{rozanecqa1:inreview, rozanecqa3:inreview}). To that end, we developed a prototype application, that simulated a manual revision process and collected users' feedback (see Fig. \ref{F:PHIA-HINTS}). Furthermore, cues were be provided to the users, to help them identify possible defects. To that end we explored three techniques: (a) GradCAM (\cite{selvaraju2017grad}), (b) DRAEM, and (c) the most similar labeled image. GradCAM is an Explainable Artificial Intelligence method suitable for deep learning models. It uses the gradient information to understand how strongly does each neuron activate in the last convolutional layer of the neural network. They are then combined with existing high-resolution visualizations to obtain class-discriminative guided visualizations as saliency masks. DRAEM (\cite{zavrtanik2021draem}) is a state-of-the-art method for unsupervised anomaly detection. It works by training an autoencoder on anomaly-free images and using it to threshold the difference between the input images and the autoencoder reconstruction. Finally, the most similar labeled images were retrieved considering the structural similarity index measure (\cite{wang2004image}). From the experiments performed (\cite{rozanecqa1:inreview, rozanecqa3:inreview}), we found that the best results were obtained when hinting the users with the images and labels with the closest structural similarity index. This resulted in an increased mean labeling time (by 30\%), but a higher quality of labeling (three times the original labeling precision, and two times the original F1 score). In addition, the number of unidentified defects was reduced by more than 80\%. Future work will explore how users' feedback can lead to discovering new defects and whether users' fatigue can be detected to alternate types of work or suggest breaks to the operators, enhancing their work experience and the quality of the outcomes.

While for this particular use case we succeeded on gathering a good quality dataset of labeled images, the process has not been straightforward on similar use cases, where multiple iterations were required, to ensure enough and high-quality data was provided. Furthermore, much work was invested towards getting few people to perform a set of lengthy experiments that provided new insights on which cues helped best towards enhancing the quality of labeling. Nevertheless, their data provided valuable insights and findings, and ground towards new directions of research.

\subsection{Human Behaviour Prediction and Safe Zone Detection}

\begin{figure*}[!t]
\centering
\includegraphics[width=0.8\textwidth]{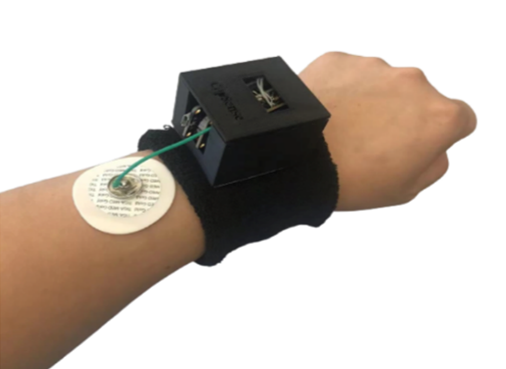}
\caption{
Caption: The sensing prototype for worker's activity recognition worn on the wrist.
Alt Text: Image of a prototype device for sensing worker's activity.
}
\label{F:DFKI-PROTOTYPE}
\end{figure*}

We have designed a sensing prototype and simple neural networks to evaluate our human activity recognition module to predict the user's movement trajectory based on artificial intelligence models. The sensing prototype (see Fig. \cref{F:DFKI-PROTOTYPE}) combines three boards: an nRF52840 backend motherboard from Adafruit Feather, a customized human body capacitance sensing board, and a data logger board. The nRF52840 board supplies three axes of Inertial Measurement Unit (IMU) data, including acceleration, gyroscope, and magnet. The customized body capacitance board is verified efficiently to sense both the body movement (\cite{bian2019passive}) and the environmental context (\cite{bian2019wrist}) by measuring the skin potential signal. The sensor data is stored in a Secure Digital card on the data logger board at a rate of 20Hz. Since the environment in the factory is full of 2.4G Hz wireless signals, such as WiFi and Bluetooth, to avoid data package loss, an SD card has been used to record the data locally and finally synchronized by checking some predefined actions. The sensing component, the IMU and body capacitance sensor, consumes the power with a level of sub-mW. A 3.7V chargeable lithium battery is used for the power supply. The feasibility of developing machine learning models for human activity recognition was validated through the development of an adversarial encoder-decoder structure with maximum mean discrepancy to realign the data distribution over multiple subjects, and tested on four open datasets. \cite{suh2022adversarial} report that the results obtained outperformed state-of-the-art methods and improved generalization capabilities. The same authors also proposed TASKED (Transformer-based Adversarial learning framework for human activity recognition using wearable sensors via Self-KnowledgE Distillation), a deep learing architecture capable of learning cross-domain feature representations using adversarial learning and maximum mean discrepancy to align data distributions from multiple domains (\cite{suh2022tasked}). Future work will address the development of models for human intention recognition and how such predictions can be leveraged for safe zone detection when routing autonomous mobile robots in the manufacturing context.

\section{Conclusions}\label{S:CONCLUSION}
The increasing digitalization of the manufacturing processes has democratized the use of artificial intelligence across manufacturing. Consequently, jobs are being reshaped, fostering the development of human-machine collaboration approaches. Humans and machines have unique capabilities, which can be potentiated through a synergistic relationship. A systemic approach is required to realize such a collaboration at its fullest. Furthermore, an architecture must be devised to support it. In particular, such an architecture must consider modules related to forecasting, explainable artificial intelligence, active learning, simulated reality, decision-making, and human feedback.

We validated the feasibility of the proposed architecture through three real-world use cases (demand forecasting, quality inspection, human behavior prediction, and safe zone detection). The experiments and results obtained in each case show how artificial intelligence can be used to achieve particular goals in manufacturing. Furthermore, it confirms the interplay between the architecture modules to deliver a human-centric experience aligned with the Industry 5.0 paradigm. Nevertheless, further research work is required to hone and highlight aspects related to workers' safety and how the current architecture supports this. 

Ongoing and future work is and will be focused on three research directions. First, we are researching human intention recognition to enhance workers' safety in industrial settings using wearable technologies. Second, we will explore active learning approaches for cybersecurity to enable real-time assessment of cyberattacks and proactively act on them. Finally, we will develop machine learning and active learning approaches for human fatigue monitoring to enhance workers' well-being in manufacturing settings.

\section*{Funding}
This work was supported by the Slovenian Research Agency and the European Union’s Horizon 2020 program projects FACTLOG and STAR under grant agreements numbers H2020-869951 and H2020-956573.

This document is the property of the STAR consortium and shall not be distributed or reproduced without the formal approval of the STAR Management Committee. The content of this report reflects only the authors' view. The European Commission is not responsible for any use that may be made of the information it contains.

\section*{Data Availability Statement }
Data not available due to restrictions.

\bibliographystyle{tfcad}
\bibliography{main}

\end{document}